\documentclass[conference]{IEEEtran}
\IEEEoverridecommandlockouts

\usepackage{cite}
\usepackage{amsmath,amssymb,amsfonts}
\usepackage{graphicx}
\usepackage{textcomp}
\usepackage{xcolor}
 \usepackage{booktabs}
\usepackage{listings}
\usepackage{algorithm}
\usepackage{algpseudocode}
\usepackage{hyperref}
\def\BibTeX{{\rm B\kern-.05em{\sc i\kern-.025em b}\kern-.08em
    T\kern-.1667em\lower.7ex\hbox{E}\kern-.125emX}}

\begin{document}

\title{Handcrafted Feature Fusion for Reliable Detection of AI-Generated Images\\

}

\author{
\IEEEauthorblockN{Syed Mehedi Hasan Nirob}
\IEEEauthorblockA{\textit{Computer Science and Engineering} \\
\textit{Shahjalal University of
Science and Technology}\\
Sylhet-3114, Bangladesh \\
smh.nirob@gmail.com}
\and
\IEEEauthorblockN{Moqsadur Rahman}
\IEEEauthorblockA{Assistant Professor \\
Shahjalal University of Science and Technology\\
Sylhet-3114, Bangladesh \\
moqsad-cse@sust.edu}
\and
\IEEEauthorblockN{Shamim Ehsan}
\IEEEauthorblockA{University of Texas at El Paso\\
El Paso, TX, 79968, USA \\
sehsan@miners.utep.edu}
\and
\IEEEauthorblockN{Summit Haque}
\IEEEauthorblockA{Assistant Professor \\
Shahjalal University of Science and Technology\\
Sylhet-3114, Bangladesh \\
summit-cse@sust.edu}
}


\maketitle

\begin{abstract}
The rapid progress of generative models has enabled the creation of highly realistic synthetic images, raising concerns about authenticity and trust in digital media. Detecting such fake content reliably is an urgent challenge. While deep learning approaches dominate current literature, handcrafted features remain attractive for their interpretability, efficiency, and generalizability. In this paper, we conduct a systematic evaluation of handcrafted descriptors---including raw pixels, color histograms, Discrete Cosine Transform (DCT), Histogram of Oriented Gradients (HOG), Local Binary Patterns (LBP), Gray-Level Co-occurrence Matrix (GLCM), and wavelet features---on the CIFAKE dataset of real vs.~synthetic images. Using 50{,}000 training and 10{,}000 test samples, we benchmark seven classifiers ranging from Logistic Regression to advanced gradient-boosted ensembles (LightGBM, XGBoost, CatBoost). Results demonstrate that LightGBM consistently outperforms alternatives, achieving PR-AUC 0.9879, ROC-AUC 0.9878, F1 0.9447, and Brier score 0.0414 with mixed features, representing a strong calibration and discrimination gain over simpler descriptors. Across three configurations (baseline, advanced, mixed), performance improves monotonically, confirming that combining diverse handcrafted features yields substantial benefit. These findings highlight the continued relevance of carefully engineered features and ensemble learning in the detection of synthetic images, particularly in contexts where interpretability and computational efficiency are critical.
\end{abstract}


\section{Introduction}
The ability of generative models such as Generative Adversarial Networks (GANs) and diffusion-based architectures to synthesize highly realistic imagery has created both opportunities and risks. On the one hand, synthetic data can enrich training corpora and drive creative applications in art, design, and simulation. On the other, the proliferation of so-called “deepfakes” undermines trust in digital media, journalism, and security systems~\cite{b8}. The misuse of AI-generated images on social media platforms has already raised concerns regarding misinformation, political manipulation, and public confusion~\cite{b10} in Bangladesh, where fake or misleading visuals are widely circulated, often outpacing fact-checking mechanisms. Developing reliable, efficient, and interpretable detection methods is therefore not only of academic interest but also of practical importance in safeguarding digital trust within Bangladesh’s rapidly growing online ecosystem.

More broadly, the societal risks of synthetic visual media are increasingly recognized at a global scale. Deepfakes have been linked to potential election interference, large-scale disinformation campaigns, and reputational damage to public figures. In digital forensics and law enforcement, courts and investigators face the challenge of verifying whether photographic evidence can be trusted. Beyond politics, the misuse of synthetic content also extends to cybercrime, harassment, and the creation of non-consensual imagery, amplifying the urgency for reliable detection frameworks.

Recent years have witnessed remarkable progress in deep learning approaches for synthetic media detection. Convolutional neural networks (CNNs) and transformer-based architectures have demonstrated state-of-the-art accuracy by exploiting subtle artifacts left by generative processes~\cite{b9}. However, these models typically demand large-scale annotated datasets, heavy computational resources, and remain criticized for their ``black-box'' nature, limiting transparency and interpretability. Such limitations restrict their applicability in resource-constrained or forensics-critical settings where trust and explainability are as important as raw performance.  

In parallel, a long line of work has explored handcrafted feature descriptors as interpretable alternatives. Classical operators such as Histograms of Oriented Gradients (HOG)~\cite{b2}, Local Binary Patterns (LBP)~\cite{b3}, and Gray-Level Co-occurrence Matrices (GLCM)~\cite{b4} are known to capture edges, texture, and spatial statistics that often differ between natural and synthesized content. Frequency-domain techniques such as the Discrete Cosine Transform (DCT) and wavelet decompositions are effective for exposing spectral irregularities introduced during up-sampling or convolutional generation. These descriptors are lightweight, data-efficient, and naturally interpretable, making them attractive for complementary or stand-alone detection pipelines.  

Despite their advantages, handcrafted features are often perceived as outdated in comparison to deep learning methods. Yet, they provide unique benefits: (i) robustness under limited training data, (ii) lower computational cost, and (iii) interpretability of the underlying decision process. Importantly, handcrafted descriptors can generalize across generative model families without requiring retraining, a property that is highly desirable in a rapidly evolving landscape of image synthesis techniques.  

In this work, we revisited the potential of handcrafted descriptors for real vs.\ fake image detection. Using the CIFAKE dataset~\cite{b1}, we systematically evaluated a diverse set of features ranging from raw pixel intensities and color statistics to texture and frequency representations. Seven classifiers are benchmarked, including linear models, bagging ensembles, and gradient-boosted decision trees. Our experimental design covers three settings: (i) baseline descriptors (raw, hist, DCT), (ii) advanced descriptors (HOG, LBP, GLCM, wavelets), and (iii) a mixed setting combining all features.  

Results reveal three key findings: (1) performance improves monotonically with richer feature sets, (2) boosted tree models (LightGBM, XGBoost, CatBoost, HistGradientBoosting) consistently outperform linear and bagging ensembles, and (3) LightGBM achieves the strongest calibration and discrimination, reaching PR-AUC 0.9879 and F1 0.9447 with mixed features. To our knowledge, this is among the most comprehensive evaluations of handcrafted descriptors on CIFAKE. Our study highlights the continued relevance of feature-engineered pipelines for robust fake image detection, especially in resource-constrained or interpretability-critical contexts such as Bangladesh, where combating misinformation has become a national priority.

\section{Related Work}

The problem of distinguishing real from synthetic images has drawn increasing research attention in recent years. Several surveys provide a comprehensive overview of detection methods, datasets, and challenges. Verdoliva~\cite{b8} reviewed early deepfake detection approaches, while Tolosana et al.~\cite{b10} provided a taxonomy of face forgery generation and detection methods. Mirsky and Lee~\cite{b11} surveyed advances in detection under the threat of ever more realistic diffusion models.

\subsection{Deep Learning Approaches}
Convolutional neural networks (CNNs) have been widely applied to detect visual artifacts in synthetic media. Li and Lyu~\cite{b9} proposed detecting warping artifacts as an early indicator of face manipulation. Masi et al.~\cite{b12} exploited mesoscopic CNN architectures for efficient deepfake detection. Attention-based models and transformers have also been explored. Zhao et al. ~\cite{b13} proposed multi-attentional features, while Wang et al.~\cite{b14} introduced a transformer-based detector robust to unseen generative models. Frequency-domain cues are another promising direction; Durall et al.~\cite{b15} showed that CNNs can leverage spectral fingerprints left by generative models.

Despite their effectiveness, CNN-based approaches face notable limitations in the context of deepfake detection. Many models exhibit poor generalization to unseen generative techniques, often overfitting to dataset-specific artifacts rather than learning robust manipulation cues~\cite{b16,b17}. This domain dependency makes them vulnerable when evaluated on cross-dataset or real-world forgeries~\cite{b18}. Furthermore, CNNs primarily capture spatial correlations and may struggle with temporal inconsistencies across video frames, which are often crucial for identifying subtle manipulations~\cite{b19}. Their reliance on high-quality input is another drawback, as performance tends to degrade significantly under compression, noise, or resolution changes, conditions common in social media environments~\cite{b20}. These challenges have motivated the exploration of architectures such as transformers and frequency-domain analysis, which aim to overcome the inherent limitations of CNN-based detectors.

\subsection{Handcrafted and Hybrid Approaches}
In parallel, researchers have investigated handcrafted or hybrid descriptors that capture statistical and textural irregularities. Operators such as Local Binary Patterns (LBP)~\cite{b3}, Gray-Level Cooccurrence Matrices (GLCM)~\cite{b4}, and HOG~\cite{b2} have been employed to capture spatial inconsistencies. Matern et al.~\cite{b16} exploited frequency and noise patterns to detect GAN images. Zhang et al.~\cite{b17} demonstrated the value of color moment statistics and steganalysis-inspired features. Although such methods may underperform deep CNNs on large benchmarks, they are lightweight, interpretable, and sometimes more robust to distribution shifts~\cite{b18}.

\subsection{Positioning of This Work}
Unlike most previous work, which emphasizes deep CNN-based detectors, our study systematically benchmarks a wide range of handcrafted descriptors in the CIFAKE dataset~\cite{b1}. By evaluating multiple feature families and classifiers, we provide quantitative evidence of their effectiveness and highlight the competitiveness of tree-based ensembles such as LightGBM~\cite{b6}, XGBoost~\cite{b5}, and CatBoost~\cite{b7}. This complements the deep learning literature by showing that interpretable and efficient hand-crafted pipelines can still achieve strong detection performance.

\section{Problem Description}
We formulate real-vs-fake image detection as a supervised binary classification task. Given an input image
\begin{equation}
x \in \mathbb{R}^{H \times W \times 3} 
\end{equation}
with label
\begin{equation}
y \in \{0,1\}, \quad 
y = \begin{cases}
0, & \text{real image},\\
1, & \text{synthetic image}.
\end{cases}
\end{equation}
the goal is to learn a classifier
\begin{equation}
f: \mathbb{R}^{H \times W \times 3} \;\rightarrow\; [0,1]
\end{equation}
that outputs a probability $p = f(x)$ of being synthetic.  

A decision is obtained via thresholding:
\begin{equation}
\hat{y} = \mathbb{I}\big[p \geq \tau \big]
\end{equation}
where $\tau$ is tuned on a validation set to maximize F1.  

To represent $x$, we extract handcrafted feature mappings covering spatial, frequency, and texture domains.  

\begin{equation}
\phi(x) = \big[\phi_{\text{raw}}, \phi_{\text{hist}}, \phi_{\text{DCT}}, 
\phi_{\text{HOG}}, \phi_{\text{LBP}}, \phi_{\text{GLCM}}, \phi_{\text{wav}}\big] 
\in \mathbb{R}^{d}
\end{equation}

The learning task then reduces to training a classifier $g$ over features:
\begin{equation}
g: \mathbb{R}^{d} \;\rightarrow\; [0,1], \qquad f(x) = g(\phi(x)).
\end{equation}

Performance is assessed using discrimination metrics (F1, ROC--AUC, PR--AUC, MCC) and calibration metrics (Brier score), ensuring both accuracy and reliability of predictions.

\section{Methodology}

\subsection{Dataset and Split}
\begin{figure}[t]
\centering
\includegraphics[width=\columnwidth]{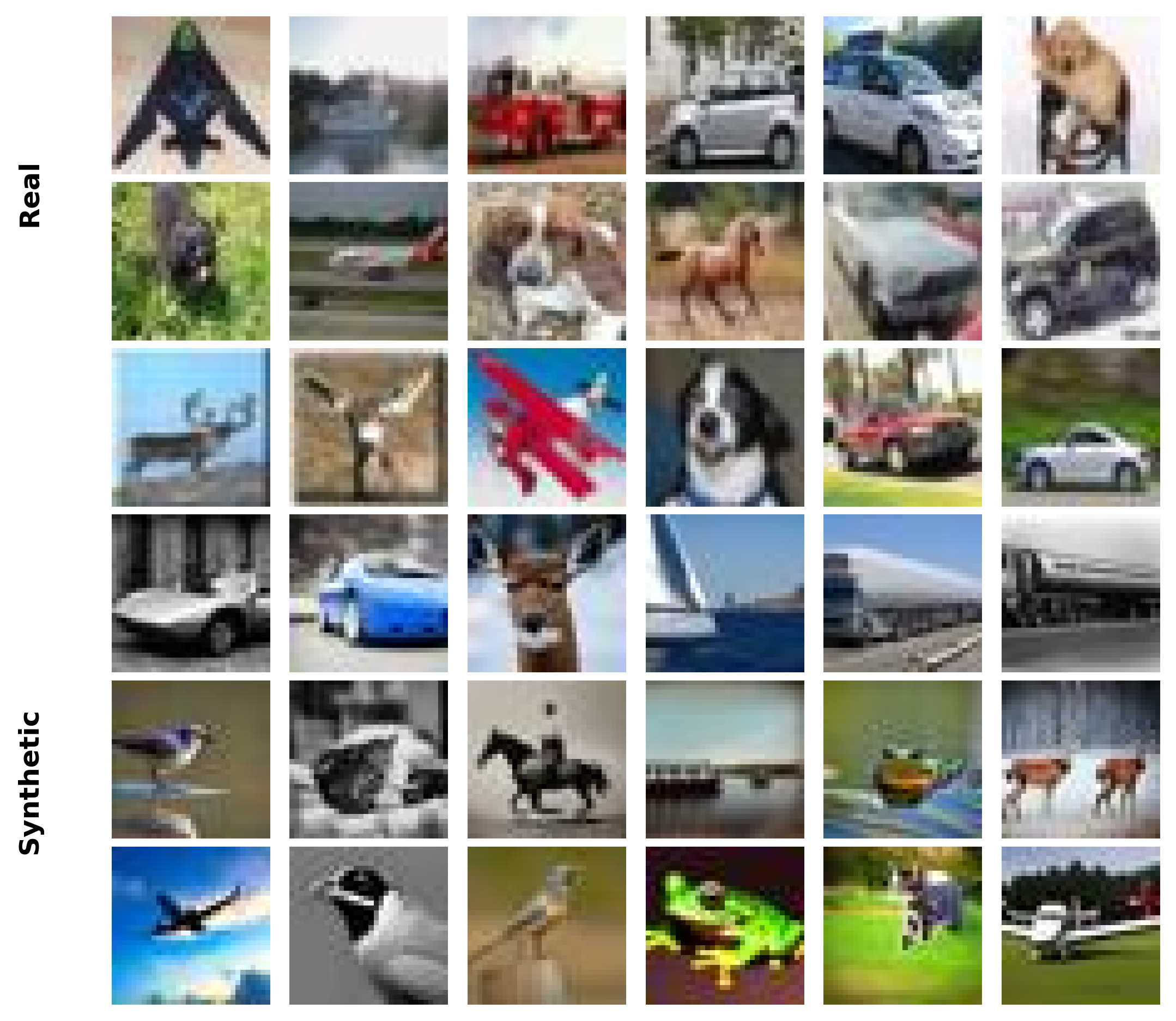}
\caption{Sample images from the CIFAKE dataset. Top three rows: Real images. Bottom three rows: Synthetic or AI-generated images.}
\label{fig:cifake_grid}
\end{figure}

We used the CIFAKE dataset~\cite{b1}, a benchmark specifically designed for real vs. synthetic image detection. The dataset consists of 120,000 images, evenly divided between authentic photographs and AI-generated counterparts. Each image is standardized to a resolution of $32 \times 32$ pixels, enabling efficient feature extraction while retaining sufficient visual detail for discrimination. Figure \ref{fig:cifake_grid} displays exemplary images from each of the two image groups.

Following our experimental protocol, we allocated 50,000 images for training and 10,000 for testing from both groups of images. Unless otherwise stated, class priors are preserved. From the training portion, we further held out a 10\% validation split via stratified sampling to tune the decision threshold (Sec.~\ref{sec:thresholding}).  

CIFAKE is chosen because it provides a balanced, controlled setting with consistent image dimensions, ensuring fair comparison across models. Unlike generic image datasets, it explicitly targets the real–synthetic boundary, making it well-suited for evaluating handcrafted features against modern generative outputs.

\subsection{Feature Extraction}
Let $x \in \mathbb{R}^{H \times W \times 3}$ be an input image and $\phi(\cdot)$ a feature map. We evaluated three configurations: (i) \emph{baseline} --- raw RGB pixels, color histograms, and DCT; (ii) \emph{advanced} --- HOG, LBP, GLCM, and wavelets; (iii) \emph{mixed} --- the concatenation of all descriptors. The final feature vector is
\begin{equation}
\mathbf{z} = \phi(x) 
\end{equation}
These feature families were chosen because they capture complementary cues: intensity and color statistics, frequency-domain regularities, edge and texture patterns, and spatial co-occurrence. Together, they provide both low-level appearance and higher-order structural descriptors, which are well-suited to expose artifacts introduced by generative models.

\noindent \textbf{Raw pixels (\,$\phi_{\text{raw}}$\,).} Normalized RGB intensities are flattened into a vector in $[0,1]$. This preserves global color distribution and serves as a simple baseline.  

\noindent \textbf{Color histograms (\,$\phi_{\text{hist}}$\,).} Per-channel histograms (16 bins/channel) are computed and $\ell_1$-normalized, capturing coarse chromatic statistics that may differ between real and synthetic content.  

\noindent \textbf{DCT (\,$\phi_{\text{DCT}}$\,).} A 2D DCT-II is applied independently to each channel, and only the top-left $k \times k$ low-frequency block ($k{=}8$) is retained, as generative models often exhibit artifacts in frequency space.

\noindent \textbf{HOG (\,$\phi_{\text{HOG}}$\,).} Histograms of Oriented Gradients~\cite{b2} are computed on grayscale (cell size $8{\times}8$, block size $2{\times}2$, 9 orientations, $L2$-Hys normalization). HOG emphasizes edge and shape distributions, useful for spotting geometric inconsistencies.  

\noindent \textbf{LBP (\,$\phi_{\text{LBP}}$\,).} Uniform Local Binary Patterns~\cite{b3} with $(P{=}8,R{=}1)$ are histogrammed into 16 bins and density-normalized. LBP encodes micro-texture, effective for identifying subtle generative noise:
\begin{equation}
\mathrm{LBP}_{P,R}(x_c)=\sum_{p=0}^{P-1} s(g_p - g_c)\,2^p, 
\quad s(t)=\mathbb{1}[t \ge 0],
\end{equation}
where $g_c$ is the intensity of the central pixel and $g_p$ are its neighbors.  

\noindent \textbf{GLCM (\,$\phi_{\text{GLCM}}$\,).} Grayscale is quantized into 32 levels, and gray-level co-occurrence matrices~\cite{b4} are computed at distances $d{=}1$ and angles $\{0,\tfrac{\pi}{4},\tfrac{\pi}{2},\tfrac{3\pi}{4}\}$. We extracted contrast, homogeneity, energy, and correlation from each matrix. These statistics capture higher-order dependencies missed by first-order histograms.  

\noindent \textbf{Wavelets (\,$\phi_{\text{wav}}$\,).} A level-1 2D discrete wavelet transform (Daubechies db2) is applied to grayscale. We computed sub-band energies for $\{LH,HL,HH\}$ and the mean and standard deviation of the approximation band $A$:
\begin{align}
\phi_{\text{wav}}(x) &= \big[\,\mathcal{E}(LH),\,\mathcal{E}(HL),\,\mathcal{E}(HH),\,\mu(A),\,\sigma(A)\,\big] \\
\mathcal{E}(B) &= \tfrac{1}{|B|}\sum_{b \in B} b^2
\end{align}
where $\mu(A)$ and $\sigma(A)$ denote the mean and standard deviation of $A$, respectively.

\subsection{Classifiers}
We evaluated a diverse set of probabilistic classifiers to probe how different learners exploit handcrafted signals:
\begin{itemize}
  \item Logistic Regression with feature standardization,
  \item Random Forest and ExtraTrees (bagging-style ensembles),
  \item HistGradientBoosting (sklearn), XGBoost~\cite{b5}, LightGBM~\cite{b6}, and CatBoost~\cite{b7} (boosting-style ensembles),
  \item Soft Voting ensembles over the available base models.
\end{itemize}
All models follow the hyperparameters in our released scripts (estimators $\sim$500–1000; learning rate $0.05$ for GBMs; regularization as specified). When a model exposed prediction probability, we used class-1 probabilities; otherwise, we mapped the decision function scores to $[0,1]$ via a rank-based monotone transform to preserve ordering.

\subsection{Training and Thresholding}
\label{sec:thresholding}
We trained on the training split and tuned the decision threshold $\tau$ on the validation split to maximize the F1 score. Let $\hat{p}_i \in [0,1]$ be the predicted probability for sample $i$ and
\begin{equation}
\hat{y}_i(\tau)=\mathbb{1}[\hat{p}_i \ge \tau]
\end{equation}
Precision and recall at threshold $\tau$ are
\begin{equation}
\mathrm{Prec}(\tau) = \frac{\mathrm{TP}(\tau)}{\mathrm{TP}(\tau)+\mathrm{FP}(\tau)}, \quad
\mathrm{Rec}(\tau) = \frac{\mathrm{TP}(\tau)}{\mathrm{TP}(\tau)+\mathrm{FN}(\tau)}
\end{equation}
We select
\begin{equation}
\tau^\star = \arg\max_{\tau \in [0,1]} \;\mathrm{F1}(\tau)
\end{equation}
with
\begin{equation}
\mathrm{F1}(\tau)=\frac{2\,\mathrm{Prec}(\tau)\,\mathrm{Rec}(\tau)}{\mathrm{Prec}(\tau)+\mathrm{Rec}(\tau)}
\end{equation}
The chosen $\tau^\star$ is then fixed and applied to the test split. This procedure ensures comparability across models and aligns with the deployment goal of maximizing balanced detection quality.

\subsection{Evaluation Metrics}
We reported both discrimination and calibration metrics, selected to provide a comprehensive view of classifier performance.  

\begin{itemize}
  \item ROC--AUC and PR--AUC (average precision) as threshold-free ranking metrics.  
  ROC--AUC captures the ability to separate classes under varying thresholds, while PR--AUC emphasizes performance on the positive (synthetic) class in imbalanced scenarios.  

  \item F1, balanced accuracy, and MCC at the selected threshold $\tau^\star$.  
  F1 balances precision and recall, balanced accuracy accounts for skewed class distributions, and MCC provides a single, correlation-based measure that remains reliable even when classes are imbalanced:
  \begin{align}
  \mathrm{MCC} &=
  \frac{\mathrm{TP}\cdot\mathrm{TN}-\mathrm{FP}\cdot\mathrm{FN}}
       {\sqrt{(\mathrm{TP}+\mathrm{FP})(\mathrm{TP}+\mathrm{FN})}} \notag \\
  &\quad \times \frac{1}{\sqrt{(\mathrm{TN}+\mathrm{FP})(\mathrm{TN}+\mathrm{FN})}}
  \end{align}

  \item Brier score to assess probability calibration:  
  \begin{equation}
  \mathrm{Brier}=\frac{1}{N}\sum_{i=1}^{N} \big(\hat{p}_i - y_i\big)^2
  \end{equation}
  This directly measures the accuracy of predicted probabilities, ensuring the model’s confidence aligns with empirical outcomes.
\end{itemize}

\subsection{Implementation Details and Reproducibility}
Feature extraction uses \texttt{Pillow}, \texttt{scikit-image} (HOG, LBP, GLCM), and \texttt{PyWavelets}. We cached per-setting features to \texttt{.npz} files to avoid recomputation. Training employs \texttt{scikit-learn} pipelines, with optional \texttt{XGBoost}, \texttt{LightGBM}, and \texttt{CatBoost} if installed. We fixed the random seed to 42 and used the stratified shuffle split for the 90/10 train/validation split. For ensembles, we used soft voting over the available probabilistic base learners. All artifacts (model, tuned threshold, metrics) are persisted for auditability.

\section{Results}\label{FAT}
We evaluated three categories of handcrafted feature sets—baseline, advanced, and mixed—on the CIFAKE dataset (50k training, 10k testing). Models are compared across six metrics: PR-AUC, ROC-AUC, F1, MCC, balanced accuracy, and Brier score. Thresholds were tuned on a held-out validation set to maximize F1.  

\begin{figure}[b]
\centering
\includegraphics[width=\columnwidth]{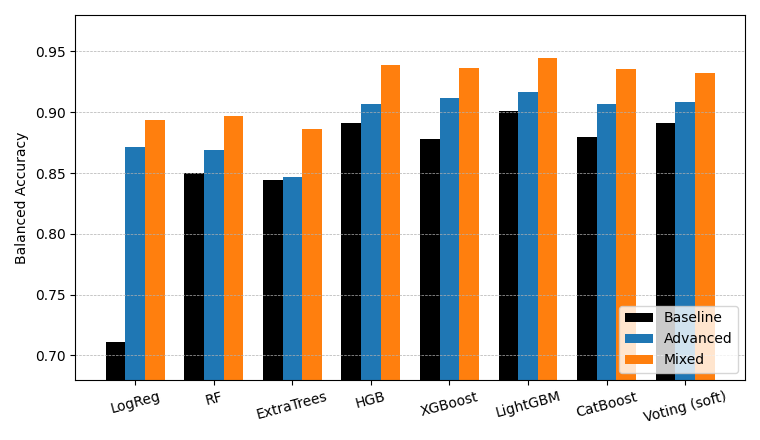}
\caption{Balanced accuracy of different models across three feature sets (Baseline, Advanced, Mixed). Accuracy improves consistently with richer feature sets.}
\label{fig:models_vs_balacc}
\end{figure}

\paragraph{Baseline descriptors} 
Our initial benchmark employed simple representations comprising raw pixels, color histograms, and DCT coefficients. As shown in Table~\ref{tab:baseline}, gradient-boosted ensembles clearly dominate. LightGBM provides the highest overall discrimination (PR-AUC 0.9676, ROC-AUC 0.9666) and calibration (Brier 0.0712), coupled with strong F1 (0.9018) and MCC (0.8022). HistGradientBoosting and XGBoost follow closely, while CatBoost also performs competitively. In contrast, bagging ensembles (Random Forest, ExtraTrees) are weaker, and Logistic Regression underperforms markedly, indicating that non-linear descriptors are best exploited by boosting-based classifiers.

\begin{table}[t]
\caption{Baseline features (raw + hist + DCT) on CIFAKE test set. 
Metrics computed at a validation-tuned threshold. Best per column in bold.}
\centering
\resizebox{\columnwidth}{!}{%
\begin{tabular}{lcccccc}
\toprule
\textbf{Model} & \textbf{PR-AUC} & \textbf{ROC-AUC} & \textbf{F1} & \textbf{MCC} & \textbf{BalAcc} & \textbf{Brier} \\
\midrule
LogReg       & 0.8057 & 0.8099 & 0.7500 & 0.4444 & 0.7115 & 0.1781 \\
RF           & 0.9270 & 0.9269 & 0.8530 & 0.7007 & 0.8501 & 0.1305 \\
ExtraTrees   & 0.9217 & 0.9232 & 0.8504 & 0.6907 & 0.8442 & 0.1372 \\
HGB          & 0.9610 & 0.9598 & 0.8916 & 0.7831 & 0.8916 & 0.0791 \\
XGBoost      & 0.9545 & 0.9530 & 0.8810 & 0.7568 & 0.8779 & 0.0868 \\
\textbf{LightGBM}     & \textbf{0.9676} & \textbf{0.9666} & \textbf{0.9018} & \textbf{0.8022} & \textbf{0.9011} & \textbf{0.0712} \\
CatBoost     & 0.9552 & 0.9536 & 0.8821 & 0.7603 & 0.8799 & 0.0870 \\
Voting (soft)& 0.9587 & 0.9579 & 0.8927 & 0.7832 & 0.8915 & 0.0925 \\
\bottomrule
\end{tabular}}
\label{tab:baseline}
\end{table}

\paragraph{Advanced descriptors} 
Incorporating structural and texture-based features (HOG, LBP, GLCM, wavelets) yields consistent gains across models (Table~\ref{tab:advanced}). LightGBM again achieves the highest performance with PR-AUC 0.9754, F1 0.9171, MCC 0.8329, and the lowest Brier score (0.0608). Compared to its baseline setting, this corresponds to a 1.5 increase in F1 and improved calibration (Brier ↓ 0.010). XGBoost, CatBoost, and HGB also see clear benefits, each crossing 0.91 F1. These results confirm that handcrafted texture and frequency descriptors expose artifacts that raw or histogram features cannot capture.

\begin{table}[t]
\caption{Advanced features (HOG + LBP + GLCM + Wavelet) on CIFAKE test set. 
Thresholds chosen on validation to maximize F1. Best per column in bold.}
\centering
\resizebox{\columnwidth}{!}{%
\begin{tabular}{lcccccc}
\toprule
\textbf{Model} & \textbf{PR-AUC} & \textbf{ROC-AUC} & \textbf{F1} & \textbf{MCC} & \textbf{BalAcc} & \textbf{Brier} \\
\midrule
LogReg        & 0.9354 & 0.9414 & 0.8742 & 0.7441 & 0.8718 & 0.0948 \\
Random Forest & 0.9447 & 0.9429 & 0.8689 & 0.7372 & 0.8686 & 0.1126 \\
ExtraTrees    & 0.9272 & 0.9263 & 0.8503 & 0.6946 & 0.8470 & 0.1356 \\
HGB           & 0.9720 & 0.9716 & 0.9089 & 0.8145 & 0.9069 & 0.0653 \\
XGBoost       & 0.9730 & 0.9728 & 0.9123 & 0.8229 & 0.9114 & 0.0637 \\
\textbf{LightGBM}      & \textbf{0.9754} & \textbf{0.9751} & \textbf{0.9171} & \textbf{0.8329} & \textbf{0.9164} & \textbf{0.0608} \\
CatBoost      & 0.9706 & 0.9708 & 0.9080 & 0.8142 & 0.9070 & 0.0668 \\
Voting (soft) & 0.9706 & 0.9709 & 0.9108 & 0.8176 & 0.9081 & 0.0715 \\
Stacking      & 0.9731 & 0.9735 & 0.9140 & 0.8247 & 0.9119 & 0.0634 \\
\bottomrule
\end{tabular}}
\label{tab:advanced}
\end{table}

\paragraph{Mixed descriptors} 
Finally, combining all descriptors (raw, hist, DCT, HOG, LBP, GLCM, wavelets) produces the strongest overall performance (Table~\ref{tab:mixed_all}). LightGBM again leads with PR-AUC 0.9879, ROC-AUC 0.9878, F1 0.9447, MCC 0.8891, and Brier 0.0414, marking a substantial jump over the advanced setting (2.8 F1 points, Brier ↓ 0.019). Other boosted trees, including XGBoost, CatBoost, and HGB, also achieve F1 scores above 0.93. Even Logistic Regression improves to F1 0.8968, though boosted ensembles remain clearly superior.  

\begin{table}[t]
\caption{Mixed features (raw + hist + DCT + HOG + LBP + GLCM + Wavelet) on CIFAKE test set. 
Thresholds chosen on validation to maximize F1. Best per column in bold.}
\centering
\resizebox{\columnwidth}{!}{%
\begin{tabular}{lcccccc}
\toprule
\textbf{Model} & \textbf{PR-AUC} & \textbf{ROC-AUC} & \textbf{F1} & \textbf{MCC} & \textbf{BalAcc} & \textbf{Brier} \\
\midrule
LogReg        & 0.9571 & 0.9600 & 0.8968 & 0.7884 & 0.8933 & 0.0773 \\
RF            & 0.9620 & 0.9611 & 0.8963 & 0.7936 & 0.8968 & 0.1026 \\
ExtraTrees    & 0.9526 & 0.9535 & 0.8884 & 0.7735 & 0.8866 & 0.1175 \\
HGB           & 0.9862 & 0.9861 & 0.9397 & 0.8783 & 0.9391 & 0.0442 \\
XGBoost       & 0.9849 & 0.9846 & 0.9370 & 0.8731 & 0.9365 & 0.0469 \\
\textbf{LightGBM}      & \textbf{0.9879} & \textbf{0.9878} & \textbf{0.9447} & \textbf{0.8891} & \textbf{0.9446} & \textbf{0.0414} \\
CatBoost      & 0.9836 & 0.9836 & 0.9358 & 0.8710 & 0.9355 & 0.0491 \\
Voting (soft) & 0.9806 & 0.9804 & 0.9329 & 0.8643 & 0.9320 & 0.0680 \\
\bottomrule
\end{tabular}}
\label{tab:mixed_all}
\end{table}

\begin{figure}[htbp]
\centering
\includegraphics[width=\columnwidth]{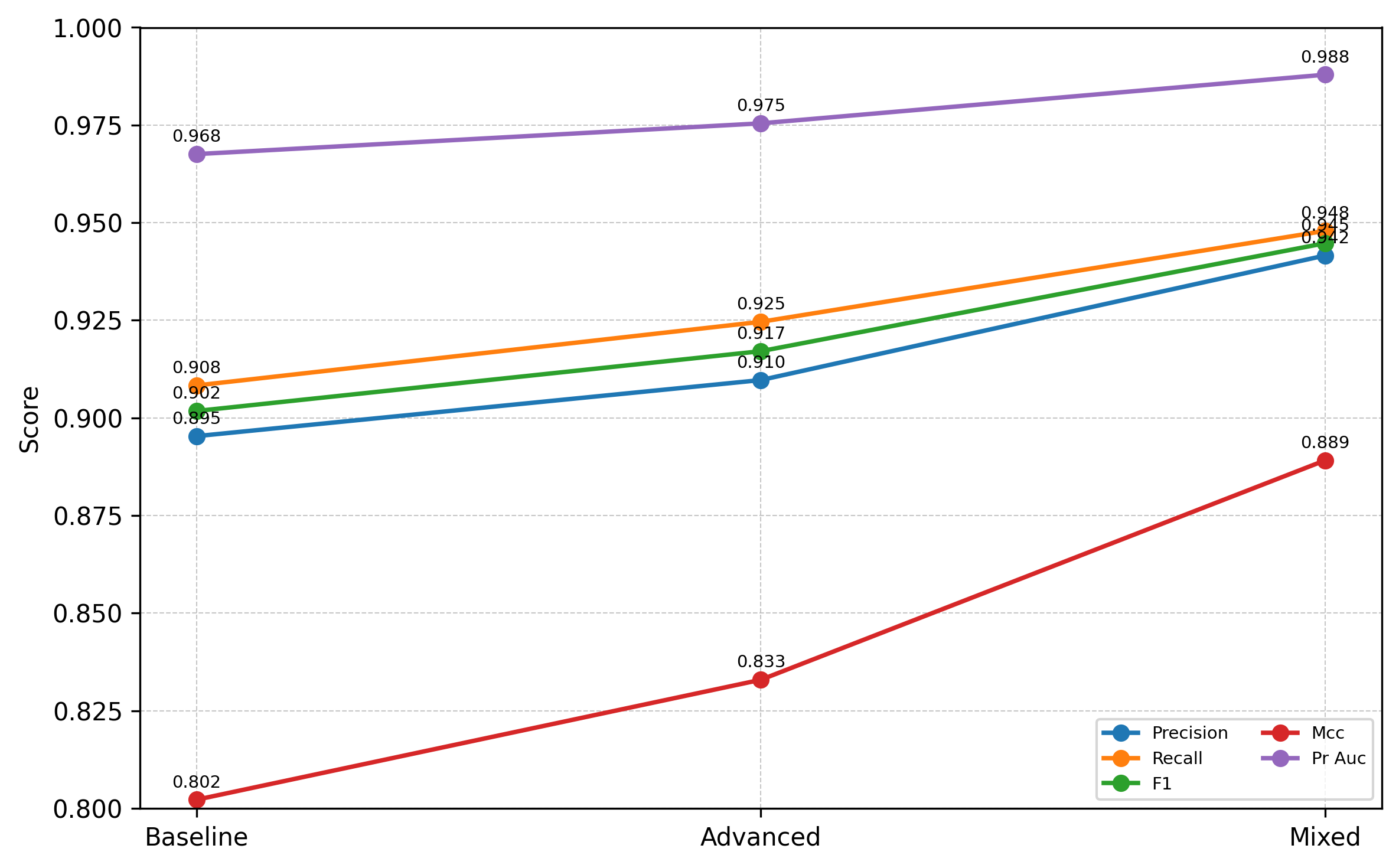}
\caption{LightGBM performance across different feature sets. All metrics improve monotonically from baseline to mixed descriptors.}
\label{fig:lgbm_metrics_by_regime}
\end{figure}

\paragraph{Cross-model comparison} 
Figure~\ref{fig:models_vs_balacc} compares balanced accuracy across models and feature sets, highlighting a consistent monotonic trend: performance improves from baseline to advanced to mixed descriptors for all classifiers. Figure~\ref{fig:lgbm_metrics_by_regime} further details LightGBM’s trajectory, showing steady gains across different metrics, including F1, PR-AUC, precision, recall, and MCC. These results reinforce that richer handcrafted features, when paired with gradient-boosted trees, provide the most reliable framework for synthetic image detection.  

\section{Discussion} 
Taken together, these experiments reveal a consistent pattern: (1) performance improves monotonically as richer feature sets are used (baseline → advanced → mixed), and (2) gradient-boosted trees (LightGBM, XGBoost, CatBoost, HGB) consistently outperform both linear (LogReg) and classical ensemble methods. LightGBM emerges as the most effective and well-calibrated classifier in all settings. 

At the same time, several limitations must be acknowledged. First, our experiments were conducted solely on the CIFAKE dataset, which—while balanced and useful—may not fully capture the diversity of generative models and artifacts found in real-world scenarios. Second, the relatively low resolution of CIFAKE images (32×32) constrains the evaluation of features that benefit from richer spatial detail. Third, we restricted our analysis to handcrafted features; hybrid pipelines that integrate handcrafted descriptors with deep representations remain unexplored. Finally, we evaluated under controlled splits (50k train, 10k test) but did not study robustness across domains, adversarial attacks, or cross-dataset generalization. 

These limitations also point to natural directions for further research. A promising hypothesis is that combining handcrafted descriptors with pretrained embeddings from vision transformers or CNN backbones could yield complementary benefits, improving both robustness and interpretability. Expanding the evaluation to richer datasets—including higher-resolution images and outputs from multiple generative model families—would provide a more realistic benchmark. Another avenue is the development of lightweight deployment frameworks for real-time detection on social media platforms, which is particularly relevant in Bangladesh and other regions where misinformation spreads rapidly, but computational resources may be constrained. Addressing these challenges will help advance the creation of practical, transparent, and generalizable detectors for synthetic visual media.

\section{Conclusion}
In this paper, we investigated the potential of handcrafted feature descriptors for detecting synthetic images, using the CIFAKE dataset as a benchmark. We extracted a wide range of features, from raw pixels and histograms to HOG, LBP, GLCM, and wavelets, and benchmarked multiple classifiers across three configurations (baseline, advanced, mixed). The experiments show a clear and consistent trend: performance improves steadily with richer feature sets, and gradient-boosted ensembles, particularly LightGBM, emerge as the most effective and well-calibrated models. This confirms that even in the era of deep learning, carefully engineered features combined with strong classical models remain highly competitive.  

Our study highlights two key insights: first, that combining diverse descriptors yields consistent improvements in discrimination and calibration; and second, that lightweight, interpretable models can provide practical value in regions such as Bangladesh, where the misuse of AI-generated images is a growing concern.  

As part of future work, we plan to extend this evaluation to richer and more diverse datasets that include multiple generative model families, higher-resolution images, and cross-domain samples. We also aim to explore hybrid approaches that combine handcrafted descriptors with deep neural representations for improved robustness. Ultimately, our goal is to advance the development of efficient, transparent, and context-aware tools for mitigating the risks posed by synthetic visual media.

\bibliographystyle{IEEEtran}
\bibliography{refs}

\vspace{12pt}

\end{document}